\documentclass{article}
\usepackage[accepted]{icml2019}

\usepackage{times}
\usepackage{soul}
\usepackage{url}
\usepackage[hidelinks]{hyperref}
\usepackage[utf8]{inputenc}
\usepackage[small]{caption}
\usepackage{graphicx}
\usepackage{amsmath}
\usepackage{amsfonts}
\usepackage{booktabs}
\begin{document}

\twocolumn[
\icmltitle{Attention-Based Selective Plasticity}

\begin{icmlauthorlist}
 \icmlauthor{Soheil Kolouri}{hrl}
  \icmlauthor{Nicholas Ketz}{hrl}
\icmlauthor{Xinyun Zou}{uci}
\icmlauthor{Jeffrey Krichmar}{uci}
\icmlauthor{Praveen Pilly}{hrl}
\end{icmlauthorlist}
 
 \icmlaffiliation{hrl}{HRL Laboratories, LLC., Malibu, CA, USA}
 \icmlaffiliation{uci}{University of California Irvine, Irvine, CA, USA}

\icmlcorrespondingauthor{Soheil Kolouri}{skolouri@hrl.com}

\icmlkeywords{Lifelong learning, Overcoming catastrophic forgetting, Selective plasticity}
\vskip 0.3in

]

\printAffiliationsAndNotice{}

\begin{abstract}

Catastrophic forgetting/interference is a critical problem for lifelong learning machines, which impedes the agents from maintaining their previously learned knowledge while learning new tasks. Neural networks, in particular, suffer plenty from the catastrophic forgetting phenomenon. Recently there has been several efforts towards overcoming catastrophic forgetting in neural networks. Here, we propose a biologically inspired method toward overcoming catastrophic forgetting. Specifically, we define an attention-based selective plasticity of synapses based on the cholinergic neuromodulatory system in the brain. We define synaptic importance parameters in addition to synaptic weights and then use Hebbian learning in parallel with backpropagation algorithm to learn synaptic importances in an online and seamless manner.  We test our proposed method on benchmark tasks including the Permuted MNIST and the Split MNIST problems and show competitive performance compared to the state-of-the-art methods.
\end{abstract}

\section{Introduction}

Existing artificial neural networks including the well celebrated deep learning architectures, such as Convolutional Neural Networks (CNNs) \cite{krizhevsky2012imagenet,he2016deep} and Recurrent Neural Networks (RNNs) \cite{graves2013speech}, are uniformly plastic. In the presence of large amounts of training data and guided by a sensible loss function, the plasticity of artificial neural networks enables them to learn from the data in an end-to-end manner and often provide the state-of-the-art performance in various applications. These include object detection \cite{redmon2017yolo9000}, action recognition from videos \cite{carreira2017quo}, speech recognition \cite{saon2017english}, and language translation \cite{edunov2018understanding}, among many others. The same uniform plasticity, on the other hand, is the culprit for a phenomenon known as ``catastrophic forgetting/interference” \cite{mccloskey1989catastrophic,mcclelland1995there}, that is, a tendency to rapidly forget previously learned tasks when presented with new training data. 


\begin{figure}[t]
    \centering
    \includegraphics[width=\columnwidth]{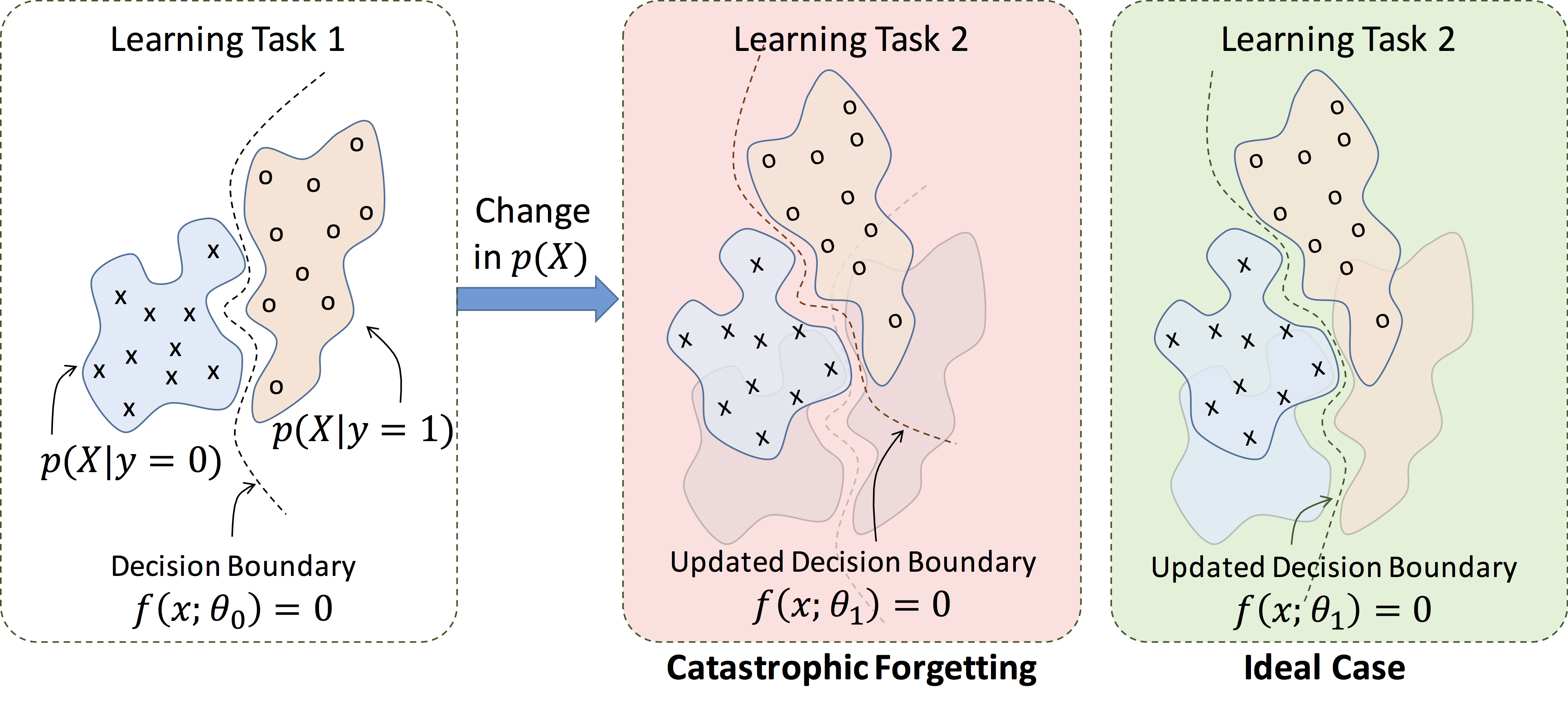}
    \caption{Depiction of catastrophic forgetting in binary classification tasks when there is a distribution shift from an initial task to a secondary task. When exposed to the distribution of the new task, the uniformly plastic parametric model, $f(\cdot,\theta)$, conforms to the new distribution with no constraints on maintaining its performance on the previous task.}
    \label{fig:intro}
\end{figure}

A uniformly plastic neural network requires independent and identically distributed samples from a stationary distribution of training samples, i.e., the i.i.d. assumption. In other words,  The `identically distributed' part of the assumption, however, is easily violated in real-world applications, specially in the continual, sequential, and lifelong learning settings. The training data could violate the identically distributed assumption, i.e., have non-stationary data distribution, when: 1) there is a shift in the distribution of the training data over time (e.g., the visual input data to a lifelong learning agent during `day' versus `night'), and 
2) the training data is not fully observable at once and different modes of variations of the data will be explored or revealed through time. This leads to a fundamental challenge in lifelong learning known as `catastrophic forgetting/interference', which indicates that a learning agent forgets its previously acquired information when learning a new task. A cartoon depiction of catastrophic forgetting is depicted in Figure \ref{fig:intro}. An ideal system should provide a balance between its plasticity and stability in order to acquire new information while preserving the old one (e.g., the decision boundary in the rightmost panel in Figure \ref{fig:intro}). 



The general idea behind our approach for overcoming catastrophic forgetting is similar in essence to the work of \cite{kirkpatrick2017overcoming,zenke2017continual,aljundi2018memory}. In short, we propose to selectively and dynamically modulate the plasticity of the synapses that are `important' for solving old tasks. Inspired by human visual cortex, we define an attention-based synaptic importance that leverages Hebbian learning \cite{hebb1961organization}. Our method is biologically inspired, in that it borrows ideas from the neuromodulatory systems in the human brain. Neuromodulators are important contributors for attention and goal-driven perception. In particular, the cholinergic system drives bottom-up, stimulus-driven attention, as well as top-down, goal-directed attention \cite{avery2014}. Furthermore, it increases attention to task-relevant stimuli, while decreasing attention to the distractions \cite{oros2014learning}. This is a similar idea to contrastive Excitatory Backpropagation (c-EB) where a top-down excitation mask increments attention to the target features and an inhibitory mask decrements attention to distractors \cite{zhang2018top}. We leverage the c-EB method and introduce a new framework for learning task-specific synaptic importance in neural networks, that enables the network to preserve its previously acquired knowledge while learning new tasks.

Our specific contributions in this work are: 
\begin{enumerate}
    \item Leveraging brain-inspired attention mechanisms for overcoming catastrophic forgetting for the first time 
    \item Hebbian learning of synaptic importance in parallel to updating synaptic weights via back-propagation and leveraging the rich literature on Hebbian learning
    \item Showing the effectiveness of the proposed method on benchmark datasets
\end{enumerate}


\section{Relevant work}

In order to overcome catastrophic forgetting three general strategies are reported in the literature: 
\begin{enumerate}
    \item selective synaptic plasticity to protect consolidated knowledge,
    \item additional neural resource allocation to learn new information, and 
    \item complementary learning for memory consolidation and experience replay.
\end{enumerate}

Interestingly, all three strategies have roots in biology. The first strategy is inspired by synaptic consolidation in the mammalian neocortex \cite{benna2016computational} where
knowledge from a previously acquired task is encoded in a subset of synapses that are rendered less plastic and therefore preserved for longer periods of time. The general idea for this strategy is to solidify and preserve synaptic parameters that are crucial for the previously learned tasks. This is often done via selective and task-specific updates of synaptic weights in a neural network. The second strategy is based on similar ideas to neurogenesis in the brain \cite{aimone2011resolving}. For a new task, allocate new neurons that utilize the shared representation learned from previous tasks but do not interfere with the old synapses. Strategy 3 is based on the theory of complementary learning systems (CLS) \cite{mcclelland1995there} in the brain and comes in various flavors. From simply recording training samples (e.g., episodic memory), to utilizing generative models (e.g., generative adversarial networks, GANs) to learn/memorize the distribution of the data.  The idea behind these methods is to make the training samples as identically distributed as possible, by adding random samples from the old distribution to the newly observed training data, providing an identically distributed data that gets close to the ideal case shown in Figure \ref{fig:intro}. 


\begin{table}[t]
\centering
\begin{tabular}{ll}
\toprule
Notation & Representing\tabularnewline
\midrule
\(f(\cdot;\theta)\) & Parametric mapping defined by a NN \tabularnewline
\(f^l_i(\cdot;\theta)\) & Output of the i'th neuron in l'th layer  \tabularnewline
\(\lambda\) & Regularization coefficient\tabularnewline
\(X\) & Input data\tabularnewline
\(x\) & Input sample\tabularnewline
\(y\) & Label\tabularnewline
\(P(\cdot)\) & Probability\tabularnewline
\(\mathcal{L}\) & Loss function\tabularnewline
\(\sigma(\cdot)\) & Nonlinearity in a neural network\tabularnewline
\(\gamma^l_{ji} \text{~~or~~} \gamma_k\) & Synaptic importance parameter  \tabularnewline
\(\theta^l_{ji} \text{~~or~~} \theta_k\) & Synaptic weights  \tabularnewline
\bottomrule
\end{tabular}
\caption{Notations used throughout the paper. }
\label{tab:notations}
\end{table}

In this paper we are interested in the first strategy, where the plasticity of synapses in a neural network are selectively and dynamically changed, allocating more plasticity to synapses that do not contribute to solving the previously learned tasks. To that end, several notable works have been recently proposed for overcoming catastrophic forgetting using selective plasticity. Some of these studies include \cite{kirkpatrick2017overcoming},  \cite{zenke2017continual}, \cite{lee2017overcoming}, and more recently \cite{aljundi2018memory}. The common theme behind all these methods is the definition of the synaptic importance parameters, $\gamma_k$, in addition to synaptic weights $\theta_k$. In all these methods, during or following learning task $A$, the synaptic importance parameters are updated along with the synaptic weights. Then, for learning task $B$, the loss function is updated to change the plasticity of different synapses with respect to their importance as: 
\begin{equation}
\mathcal{L}(\theta)=\mathcal{L}_B(\theta)+\underbrace{\lambda\sum_{k} \gamma_k(\theta_k -\theta^\star_{A,k})^2}_{\text{Regularizer}} 
\label{eq:updated_loss}
\end{equation}
where $\theta^\star_{A,k}$ are the optimized synaptic weights for task $A$, and $\mathcal{L}_B(\theta)$ is the original loss function for learning task $B$, e.g., the cross entropy loss. Intuitively, the regularizer penalizes large change for synapses that are important for solving task $A$, therefore, the network is forced to utilize synapses that are less important for previously learned tasks to solve a new one. The difference between these methods is on the way they calculate the importance parameters, $\gamma_k$. 

\begin{figure*}[t]
    \centering
    \includegraphics[width=\linewidth]{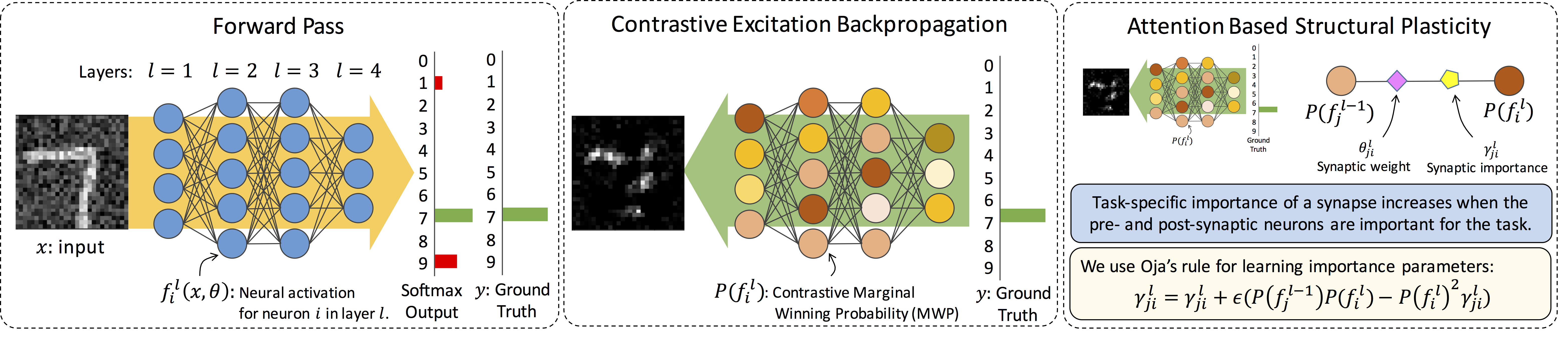}
    \caption{Illustration of our proposed framework for continual learning. Connections in the neural network are committed to a given task based on contrastive excitation backpropagation (c-EB). For each training example (e.g., an image with a ``7"), c-EB is applied with the ground truth label (``7" here) to generate attentional maps at each upstream layer in the hierarchy. A connection is considered important for a given task (e.g., classifying digits for a particular MNIST task) if its pre- and post-synaptic neurons are highlighted by the c-EB process. We use Oja's rule to incrementally update the importance of such connections during task learning. This procedure consolidates various important connections in the network for experienced tasks, preventing their forgetting as new tasks are learned.}
    \label{fig:framework}
\end{figure*}

In the Elastic Weight Consolidation (EWC) work, Kirkpatrick et al. provide a Bayesian argument that the information about task $A$ is fully absorbed in the posterior distribution $p(\theta\vert X_A)$. Then, they approximate the posterior as a Gaussian distribution with mean given by $\theta^\star_{A}$ and a diagonal precision matrix given by the Fisher information matrix, $F$, where they set the importance parameter to be the diagonal values of this matrix, $\gamma_k=F_{kk}$. The methods proposed by Kirkpatrick et al., however, is not online. In the sense that the importance parameters are calculated at the end of learning each task in an offline manner. Zenke et al. and  Aljundi et al. provided online variations of the EWC. Specifically, Zenke et al. set the synaptic importance, $\gamma_k$, to be a function of the cumulative change a synapse experiences during training on a specific task. They denote their algorithm as Synaptic Intelligence. The more cumulative changes correspond to more importance. Similarly, Aljundi et al. consider the importance as the cumulative effect of a synapse on the norm of the last layer of the neural network before the softmax classifier, hence decoupling the importance parameters from labels and enabling the importance parameters to continue to update even in absence of labels. Aljundi et al. further show that their proposed importance is equivalent to calculating the Hebbian trace of a synapse. 
In this paper, we follow the existing work in the literature, but bring in a biologically plausible solution based on neuromodulatory attentional mechanisms in the human brain. 

\section{Method}
Our proposed method leverages the bio-inspired top-down attention mechanism of contrastive excitation backpropagation (c-EB), to update synaptic importance parameters of a network in an online fashion. Figure \ref{fig:framework} depicts the core idea in our proposed framework. We denote the notations used throughout this paper in Table \ref{tab:notations}.



\subsection{Excitation back-propagation}

Excitation Back-Propagation and its contrastive variation are biologically inspired top-down attention mechanisms \cite{zhang2018top}, which are used in computer vision applications as visualization tools for CNNs' top-down attention. With an abuse of notation we let  $f^l_i$ denote the i'th neuron in layer $l$ of a neural network. Define the \emph{relative importance} of neuron $f^{(l-1)}_{j}$ on the activation of neuron $f^l_{i}$, where $f^l_{i}=\sigma(\sum_{ji}\theta^{l}_{ji}f^{(l-1)}_{j})$ and for $\theta^{l}$ being the synaptic weights between layers $(l-1)$ and $l$, as a probability distribution $P(f^{(l-1)}_{j})$ over neurons in layer $(l-1)$. This probability distribution can be factored as,
\begin{equation}
P(f^{(l-1)}_{j}) = \sum_{i}P(f^{(l-1)}_{j} \vert f^l_{i})P(f^l_{i}).
\label{eq:ebp0}
\end{equation}
$P(f^{l}_{i})$ is the Marginal Winning Probability (MWP) for neuron $f^{l}_{i}$, Zhang et al. then define the conditional probability $P(f^{(l-1)}_{j} \vert f^l_{i})$ as
\begin{equation}
P(f^{(l-1)}_{j} \vert f^l_{i}) =
\begin{cases}
Z^{(l-1)}_{i}f^{(l-1)}_{j}\theta^{l}_{ji} & \text{if } \theta^{(l-1)}_{ji} \geq 0, \\
0 & \text{otherwise},
\end{cases}
\label{eq:ebp}
\end{equation}
where  $$Z^{(l-1)}_{i}=\left(\sum_j f^{(l-1)}_{j}\theta^{l}_{ji}\right)^{-1}$$ is a normalization factor such that $\sum_{j}P(f^{(l-1)}_{j} \vert f^l_{i}) = 1$. For a given input, $x$, (e.g., an image), EB generates a heat-map in the pixel-space w.r.t. class $y$ by starting with $P(f^{L}_{i}=y)=1$ at the output layer and applying Equation (\ref{eq:ebp}) recursively. 

Furthermore, the contrastive-EB (c-EB) assigns a hypothetical negative node $\bar{f}^L_i$, with weights $\bar{\gamma}^L_{ji}=-{\gamma}^L_{ji}$.c-EB then recursively calculates  $\bar{P}(f^{(l-1)}_{j} \vert f^l_{i})$ for this negative node $\bar{f}^L_i$. The final \emph{relative importance} of the neurons is then calculated as a normalized difference of $P(f^{(l-1)}_{j} \vert f^l_{i})$ and $\bar{P}(f^{(l-1)}_{j} \vert f^l_{i})$, 
$$ P_{c}(f_j^{(l-1)}\vert f_i^{l})=\frac{ReLU(P(f^{(l-1)}_{j} \vert f^l_{i})-\bar{P}(f^{(l-1)}_{j} \vert f^l_{i}))}{\sum_j ReLU(P(f^{(l-1)}_{j} \vert f^l_{i})-\bar{P}(f^{(l-1)}_{j} \vert f^l_{i}))}$$ 
where $ReLU$ is the rectified linear function. Finally, the contrastive-MWP, $P_{c}(f_i^{l})$, indicates the relative importance of neuron $f_i^{l}$ for specific prediction $y$. Alternatively, $P_{c}(f_i^{l})$ could be thought as the implicit amount of attention that the network  pays to neuron $f_i^{l}$ to predict $y$. Next, we will use the contrastive-MWPs to update the synaptic importance parameters.

\subsection{Attention-Based Synaptic Importance}

Let $\gamma_{ji}^{l}$ denote the importance of the synapse between neurons $f_j^{(l-1)}$ and $f_i^l$ for a particular task. Here we hypothesize that the importance of a synapse should be increased if its pre and post synaptic neurons are important (relative to the task that is being learned), where the importance of the neurons are identified via Equation \eqref{eq:ebp0}. This is the basic idea behind Hebbian learning \cite{hebb1961organization}. Hebbian learning of importance parameters, however, suffers from the severe problem of unbounded growth of these parameters. To avoid the unbounded growth and following the large body of work on Hebbian learning, we use Oja's learning rule \cite{oja1982simplified} that provides an alternative and more stable learning algorithm.  We then update the importance parameters as follows:
\begin{equation}
    \gamma_{ji}^{l}=\gamma_{ji}^{l}+\epsilon\left(P_c(f^{(l-1)}_j)P_c(f^{(l)}_i)- P_c\left(f^{(l)}_i\right)^2\gamma^l_{ji}\right)
\end{equation}
where $\epsilon$ is the rate of Oja's learning rule.

While the network is being updated via back-propagation, we also update the importance parameters via Oja's learning rule in an online manner, starting from $\gamma_{ji}^l=0$. 


\begin{figure}[t]
    \centering
    \includegraphics[width=\columnwidth]{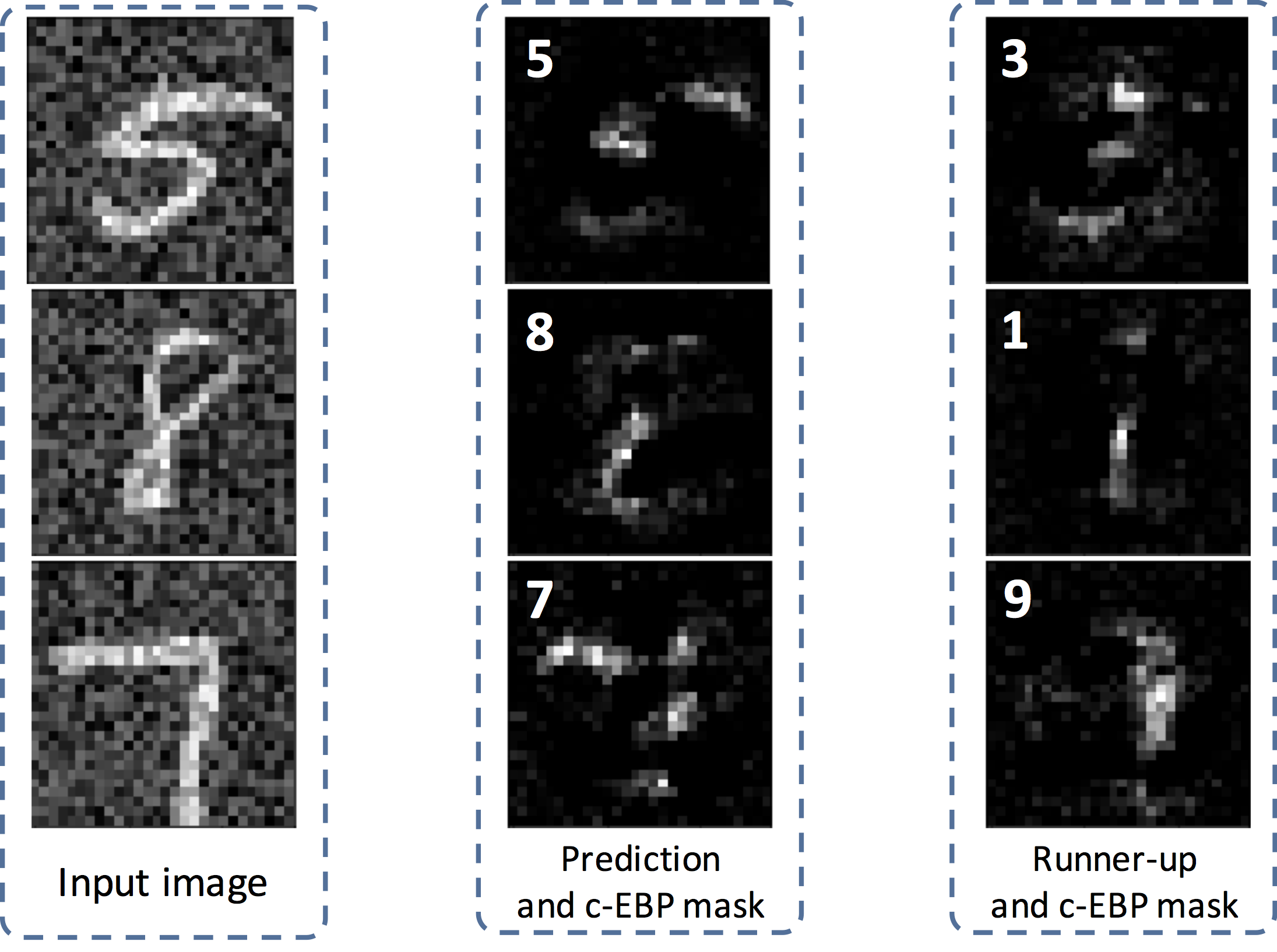}
    \caption{The visualization of c-EB at the input layer for different top-down signals. The first column shows the input image, the second column shows the attentional map generated by c-EB for the predicted label (i.e., with highest activity after the softmax layer), and the third column is for the runner-up predicted label.}
    \label{fig:cEBP}
\end{figure}

\subsection{Updated loss}

Following the existing work for overcoming catastrophic forgetting \cite{kirkpatrick2017overcoming,zenke2017continual} we regularize the loss function with the computed synaptic importance parameters as in Equation \eqref{eq:updated_loss}, i.e.,
$$\mathcal{L}(\theta)=\mathcal{L}_B(\theta)+\lambda\sum_{k} \gamma_k(\theta_k -\theta^\star_{A,k})^2 $$
We further note that, as opposed to the work of \cite{kirkpatrick2017overcoming} and similar to the work of \cite{zenke2017continual,aljundi2018memory} the importance parameters in our work could be calculated in an online fashion. Therefore, there is no need for definition of tasks, and our method could adaptively learn the changes in the training data. However, in order to be able to compare our results with those of the Elastic Weight Consolidation (EWC), we use the exact loss function used in that work \cite{kirkpatrick2017overcoming}. As can be seen the c-EB is capable of identifying parts of the input image (i.e., neurons in layer 0) that correspond to the top-down signal. 


\begin{figure}[t]
    \centering
    \includegraphics[width=\columnwidth]{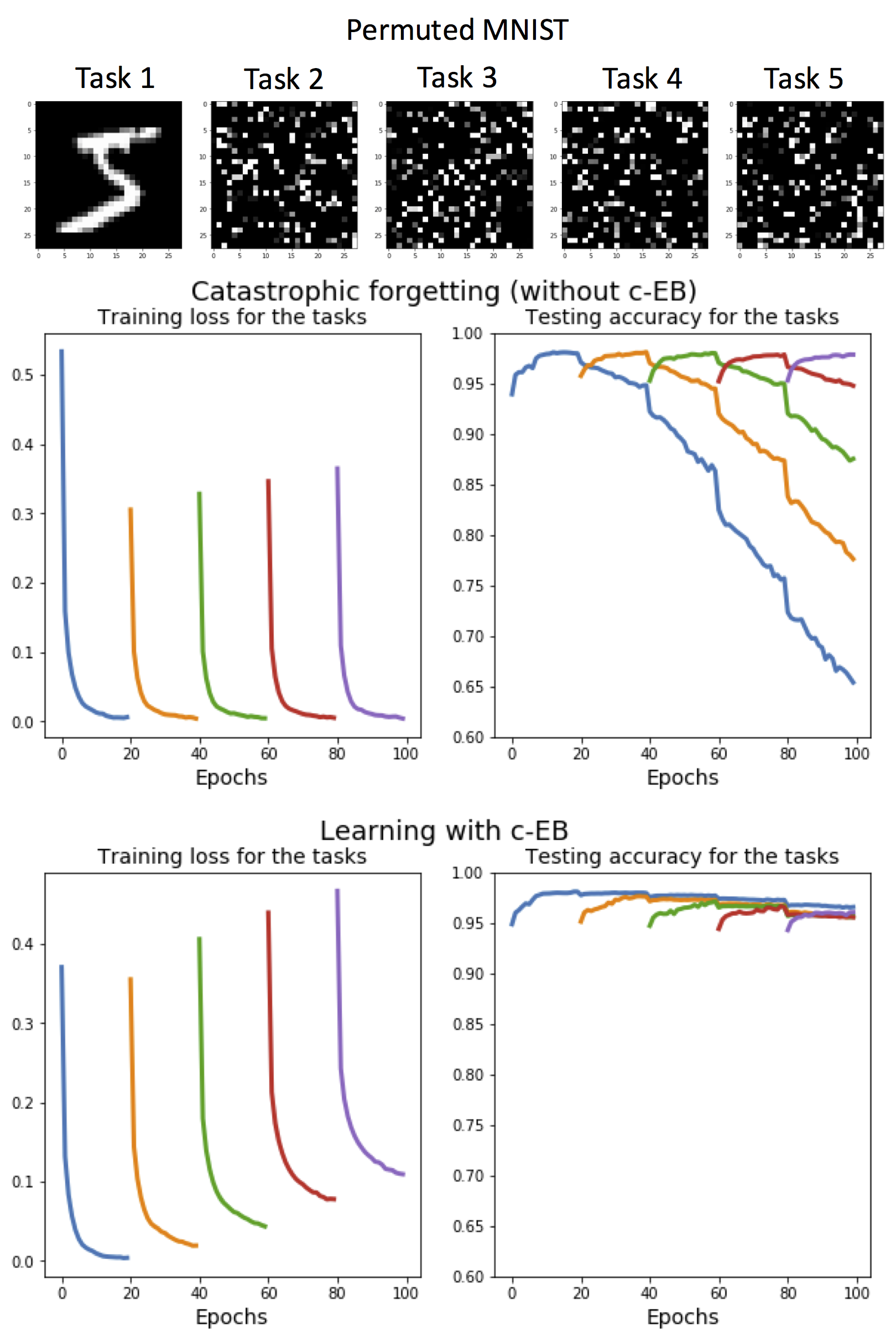}
    \caption{Performance of our algorithm on the Permuted MNIST tasks in comparison with and without c-EB.}
    \label{fig:permMNSITAcc}
\end{figure}

\section{Experiments}

\subsection{Permuted MNIST} 

We test our algorithm on the benchmark permuted MNIST task, with five sequential tasks. The first task is set as the original MNSIT problem while the consequent tasks  obtained by fixed but random permutations of the digit images (See Figure \ref{fig:permMNSITAcc}) top row). We start by learning the first task, i.e., original MNIST problem, with our attention-based selectively plastic multilayer perceptron. After training on original MNIST and achieving saturated accuracy ($\sim 98\%$), we test our c-EB top-down attention. We first add Gaussian noise to MNIST test images and calculated the attention maps at the input layer setting the top down signal to be: 1) the predicted label (i.e., neuron with the highest activation after softmax layer), and 2) the runner up predicted label (i.e., the neuron with the second highest activation). The inputs and their corresponding attention maps for three sample digits are shown  in Figure \ref{fig:cEBP}.


The result on learning the consecutive permuted MNIST problems is shown in Figure \ref{fig:permMNSITAcc}. We followed the work of \cite{kirkpatrick2017overcoming} and used a Multi-Layer Perceptron (MLP) with two hidden layers of size $400$ (each). We used Rectified Linear Units (ReLUs) as nonlinear activation functions and the ADAM optimizer with learning rate, $lr=1e-3$, for optimizing the networks. We report the average training loss as well as the average testing accuracy over 10 runs for all five tasks, for a vanilla network, i.e., a uniformly plastic neural network without selective plasticity, and for our proposed method.   It can be seen that the Vanilla network suffers from catastrophic forgetting while our attention-based selective plasticity enables the network to preserve its important synapses. 

\begin{figure}[t]
    \centering
    \includegraphics[width=\columnwidth]{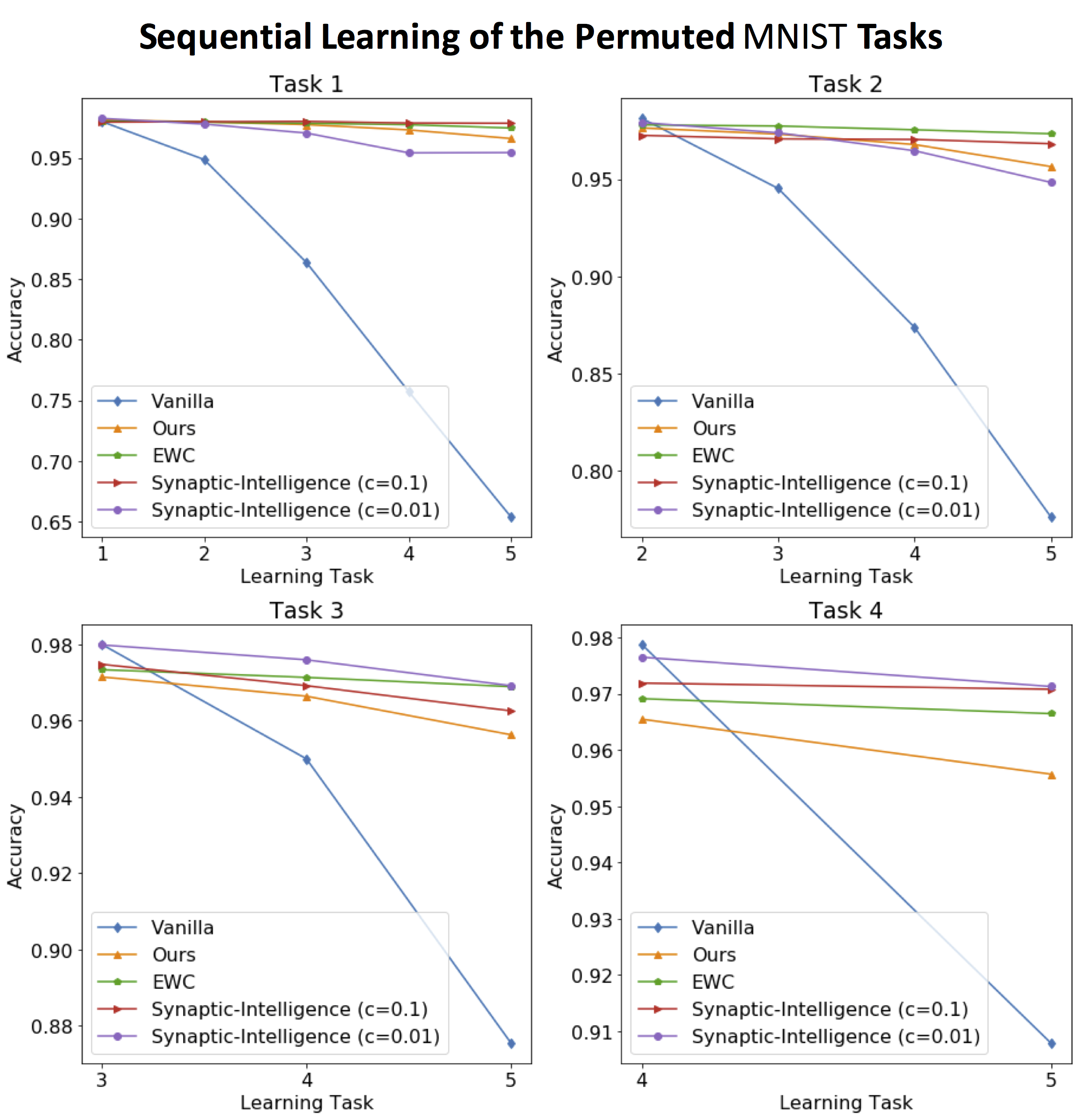}
    \caption{Comparison between our method, EWC \cite{kirkpatrick2017overcoming}, and Synaptic Intelligence \cite{zenke2017continual} (where $c$ is a hyper-parameter for Synaptic Intelligence). As can be seen our method performs on par with these algorithms. We emphasize that we spent little to no efforts on hyperparameter tuning for our algorithm. }
    \label{fig:permMNIST}
\end{figure}

Furthermore, we compared our performance to that of the EWC \cite{kirkpatrick2017overcoming} and  Synaptic Intelligence \cite{zenke2017continual}. The networks architecture, optimizer, learning rates, and batch size (batch size=100) was kept the same for all methods and we used the optimal hyper parameters reported in these papers. We emphasize that we performed little to no hyper-parameter tuning for our algorithm.  The comparison between the methods is shown in Figure \ref{fig:permMNIST}. 
Each plot in Figure \ref{fig:permMNIST} shows the classification accuracy for task $t$ after learning tasks $t,t+1,...,T=5$. An ideal system should provide high accuracy for task $t$, and maintain it when learning the subsequent tasks.  As can be seen our method performs on par with the SOA algorithms and we suspect a better hyper-parameter tuning would in fact further boost the results of our algorithm. 

\subsection{Split MNIST} 

\begin{figure}[t]
    \centering
    \includegraphics[width=\columnwidth]{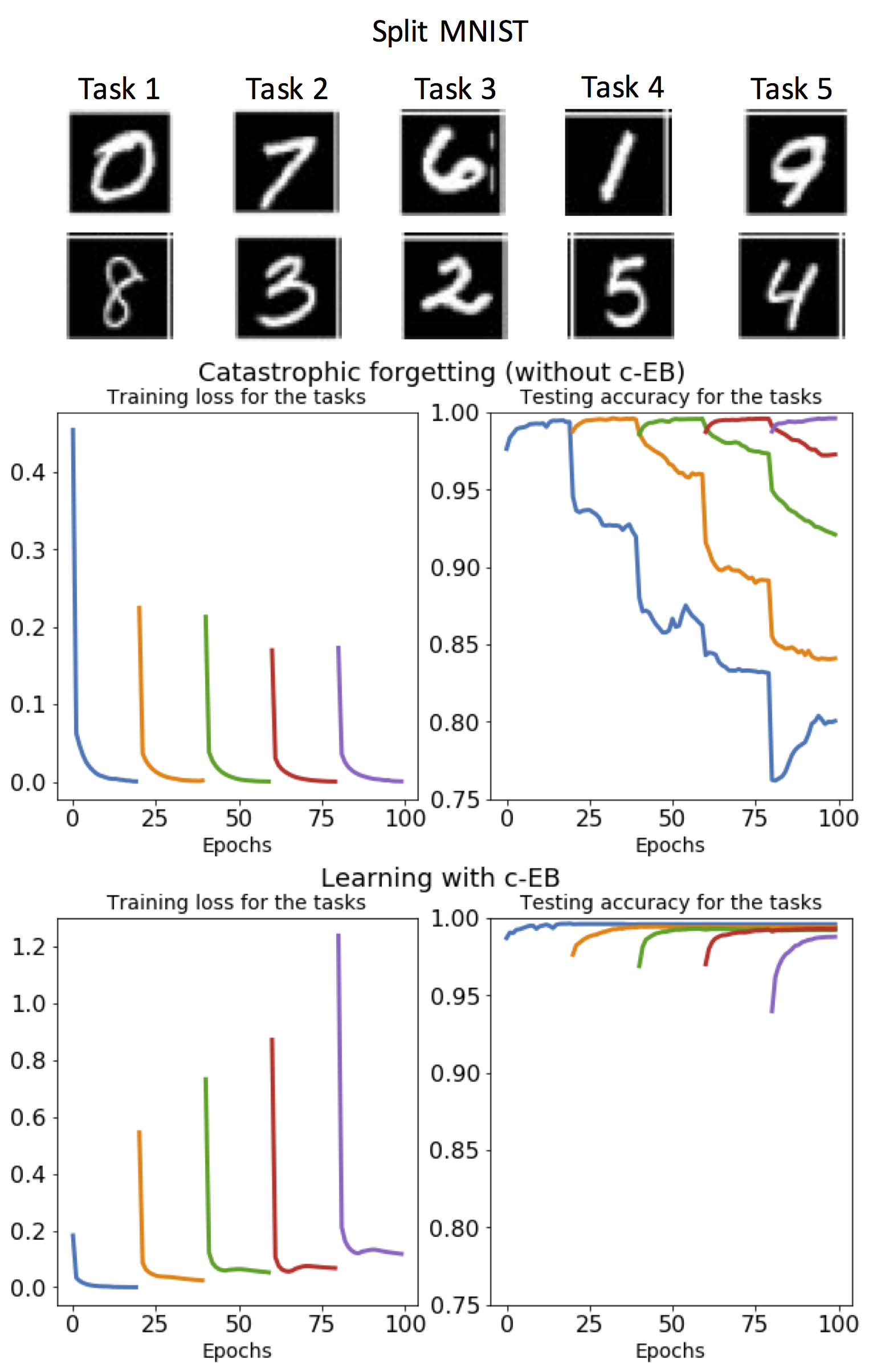}
    \caption{Performance of our algorithm on the Split MNIST tasks in comparison with and without c-EB.}
    \label{fig:splitMNSITAcc}
\end{figure}

For the Split MNIST tasks we learn five consecutive pairs of digits, (e.g., $[0,5],[1,6],[2,7],[3,8],[4,9]$) where the pairs are randomly chosen. The Split MNIST task is a more realistic lifelong learning scenario compared to the Permuted MNIST task. In Split MNIST, knowledge from the previously learned tasks could be transferred to learning future tasks. Figure \ref{fig:splitMNSITAcc} shows the performance of our algorithm on the split MNIST tasks and compare it to a vanilla neural network with the same architecture.

\begin{figure}[t]
    \centering
    \includegraphics[width=\columnwidth]{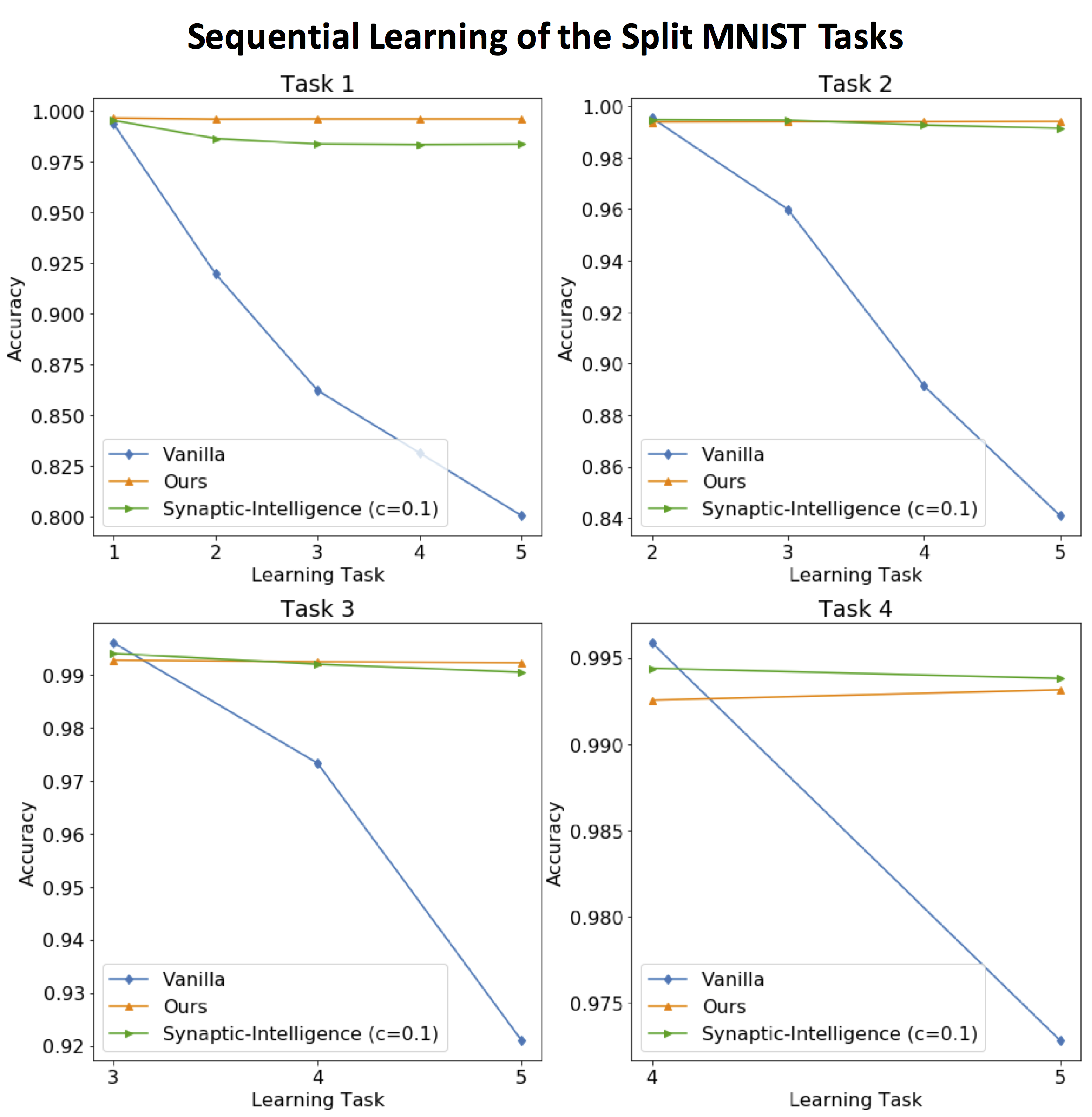}
    \caption{Comparison between our method and Synaptic Intelligence \cite{zenke2017continual} on the Split MNIST tasks. As can be seen our method performs on par with synaptic intelligence. We emphasize that we spent little to no efforts on hyper-parameter tuning for our algorithm. }
    \label{fig:splitMNIST}
\end{figure}

Finally, we compare our work with the Synaptic Intelligence \cite{zenke2017continual} on the Split MNIST tasks. The results of this comparison are shown in Figure \ref{fig:splitMNIST}.

\section{Conclusion}

In this paper we introduced a biologically inspired mechanism for overcoming catastrophic forgetting. We propose a top-down neuromodulatory mechanism for identifying important neurons relevant to the task. We then attach an importance parameter to all synapses in the neural network and update this importance based on Oja's learning rule on pre and post synaptic importance of neurons. This is a novel online method for synaptic consolidation in neural networks to preserve previously acquired knowledge. While our results were demonstrated for sequential acquisition of classification tasks, we believe the biological principle of top-down attention driven by the cholinergic neuromodulatory system would also be applicable to deep reinforcement learning networks. Future work will also look at other ways of implementing top-down attention such as Grad-CAM \cite{selvaraju2017grad}, and establish the generality of the principle. 


\bibliographystyle{named}
\bibliography{ABSP}
\end{document}